\documentclass[10pt,journal,compsoc]{IEEEtran}

\usepackage{}
\usepackage{amsmath}
\usepackage{amsthm}
\usepackage{amssymb}
\usepackage{amsfonts}
\usepackage{graphicx}
\usepackage{url}
\usepackage{subfigure}
\usepackage{float}
\usepackage{mathrsfs,fourier,pifont}
\usepackage{threeparttable}
\usepackage{algorithmic}
\usepackage{algorithm}
\usepackage{color}
\usepackage{booktabs}
\usepackage{epstopdf}
\usepackage{multicol,multienum, multirow}
\usepackage{makecell}
\usepackage{tablefootnote}
\usepackage{hyperref}

\usepackage{pgfplots}

\usepgfplotslibrary{groupplots}
\usetikzlibrary{patterns}
\usetikzlibrary{secdia}
\usepgfplotslibrary{polar}
\usepackage{pgfplotstable}
\usepackage{siunitx}
\usepackage{tikz}

\makeatletter
\def\th@plain{%
  \rm 
}
\makeatother


\begin{document}

\title{Non-iterative recomputation of dense layers for performance improvement of DCNN}

\author{Yimin~Yang, \IEEEmembership{Member,~IEEE,}
~Q.~M.~Jonathan Wu, \IEEEmembership{Senior Member,~IEEE}, Xiexing~Feng, \IEEEmembership{Graduate Student Member,~IEEE,} and Thangarajah~Akilan, \IEEEmembership{Graduate Student Member,~IEEE,}
\thanks{This work was supported by
the Natural Sciences and Engineering Research Council of Canada.
}
\thanks{Y.~M.~Yang is with the Department of Electrical and Computer Engineering, University of Windsor N9B 3P4, Canada, and also with the Computer Science Department, Lakehead University, Thunder Bay, P7B 2A4, Canada (yyang48@lakeheadu.ca)}
\thanks{Q.~M.~J. Wu, X.~X.~Feng, and T.~Akilan are with the Department of Electrical and Computer Engineering, University of Windsor N9B 3P4, Canada.}
}


\maketitle

\begin{abstract}

An iterative method of learning has become a paradigm for training deep convolutional neural networks (DCNN).  However, utilizing a non-iterative learning strategy can accelerate the training process of the DCNN and surprisingly such approach has been rarely explored by the deep learning (DL) community. It motivates this paper to introduce a non-iterative learning strategy that eliminates the backpropagation (BP) at the top dense or fully connected (FC) layers of DCNN, resulting in, lower training time and higher performance. The proposed method exploits the Moore-Penrose Inverse to pull back the current residual error to each FC layer, generating well-generalized features. Then using the recomputed features, i.e., the new generalized features the weights of each FC layer is computed according to the Moore-Penrose Inverse. We evaluate the proposed approach on six widely accepted object recognition benchmark datasets: Scene-15, CIFAR-10, CIFAR-100, SUN-397, Places365, and ImageNet. The experimental results show that the proposed method obtains significant improvements over 30 state-of-the-art methods. Interestingly, it also indicates that any DCNN with the proposed method can provide better performance than the same network with its original training based on BP.

\end{abstract}

\section{Introduction}

\begin{figure*}
  \centering
\includegraphics[width=5in,height=2.6in,origin=br]{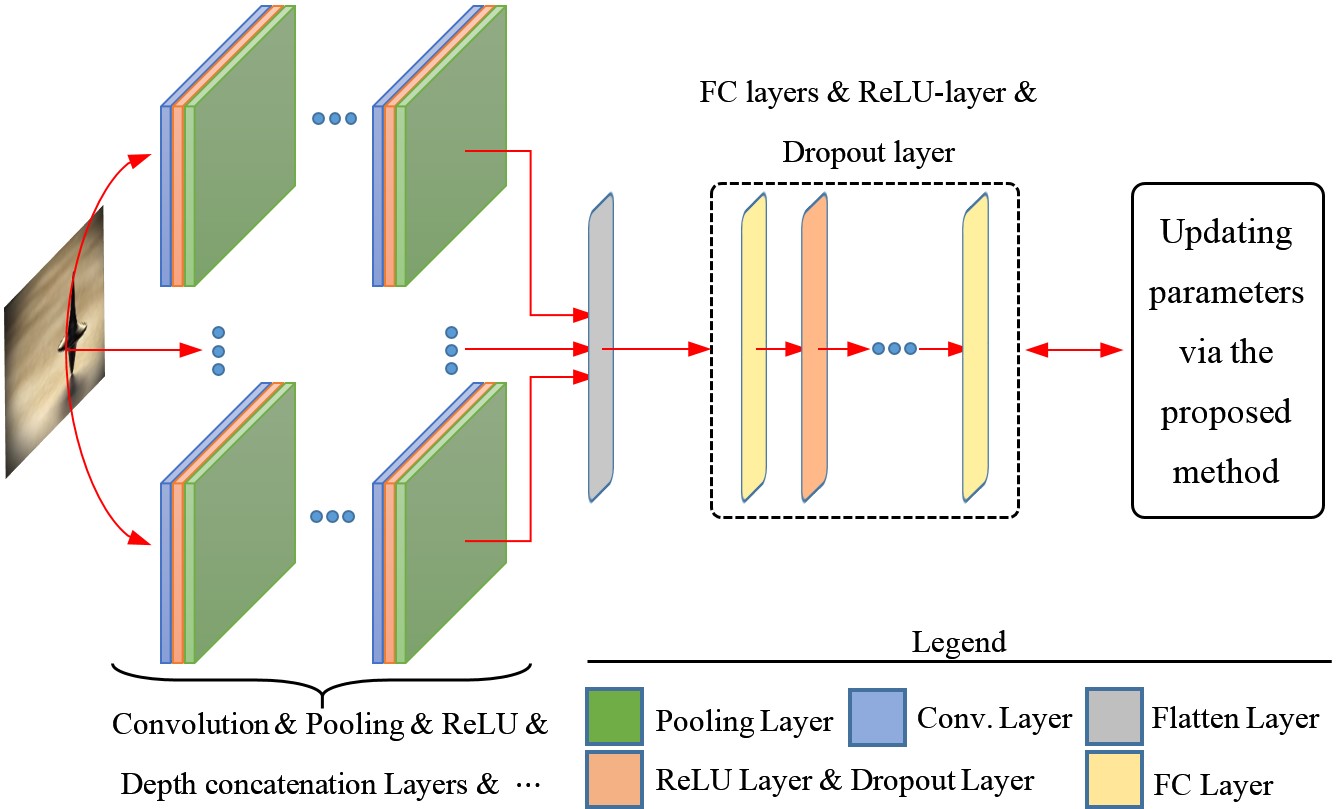}
\caption{Schematic Diagram of the proposed method}
\label{figure1}
\end{figure*}

The past few years have witnessed the bloom of DL including auto-encoders, DCNN, etc \cite{888,weng1992,LeCun:89,Hinton2006,Bengio07greedylayer-wise,Vincent2010,minminchen2015,Hintonnature2015}. DL has been around for many years dating back to the works in the 1980s \cite{888,weng1992,fukushima:1980,Bengio:94,Schraudolph:93,LeCun:89}. The \emph{Neocognitron} \cite{fukushima:1980} was probably the first network that deserved a deep structure and was the first to incorporate neurophysiological insights. Hinton \emph{et al.} initiated a breakthrough that \cite{Bengio07greedylayer-wise, Vincent2010,minminchen2015} by multilayer NN with the BP that was used to reduce the dimensionality of data. Over many benchmark datasets, recent DL methods including GoogLeNet \cite{Szegedy_2015_CVPR}, AlexNet \cite{NIPS7433451}, very deep convolutional network \cite{Chatfield14}, 96/160-layer ResNet \cite{Liang2015}, Network in Networks \cite{NINyan}, Google-Inception model \cite{7298594}, and DenseNet \cite{Huang_2017_CVPR}, have substantially advanced the state-of-the-art accuracies of objection recognition and have turned out to be very good at discovering intricate structures in real data. With NN depth increasing, the richness of the data representation is enhanced, and the generalization performance of the final classifier improves as well. Recent evidence reveals that network depth is crucial importance, as the classification/recognition results of deeper NN are better than the shallow ones. It can be seen from the DCNNs with a depth of 8-layer of of AlexNet \cite{NIPS7433451}, 16-layer of VGG \cite{Chatfield14}, and 152-layer of ResNet \cite{He_2016_CVPR}. With network depth increasing or network architecture optimization, the performance of DCNN methods has been boosted significantly and is therefore applicable to many real-world applications, like image recognition, semantic, segmentation, and so forth.

However, the performance improvements through network architecture modification is approaching its limitation according to the recent results on the ILSVRC competition. For example, compared to the 8-layer AlexNet, which is the winner of ILSVRC 2012, the winner of ILSVRC 2014 a 19-layer VGG model achieves 9.1$\%$ lesser top-5 error. However, after the year of 2015, any single modality DCNN almost maintains its performance regardless of the number of layers increases to hundreds of layers. For instance, 152-layer ResNet \cite{He_2016_CVPR}, and 316-layer Inception-v3 provide 4.49$\%$, 6.12$\%$, and 5.6$\%$ top-5 error rates respectively on the ImageNet validation set. Thus a motivation naturally comes: \emph{\textbf{Can we further improve the performance of the DCNN models by a new learning method}? }

Although a lot of research efforts has accomplished architectural improvements in the DCNN, all the present-day DCNN models use the BP as a cornerstone of their end-to-end training. Such iterative training process of BP suffers from slow convergence, getting trapped in a local minimum and being sensitive to the learning rate configurations.  Unlike iterative learning strategy, the non-iterative methods have emerged into the single-layer-based classifiers for a long time \cite{SCHMIDT1992}. Likewise, the Moore-Penrose Inverse exploited in this paper can be referred to the work of Schmid \cite{SCHMIDT1992} back in 1992. Authors in \cite{SCHMIDT1992} mentioned that the neuron weights could sometimes be called the Fisher vector and found by solving the linear equations through standard numerical methods, such as BP or the generalized inverse method. Later in 2004, Huang \emph{et al.} \cite{1Huang2004} proved that with Moore-Penrose Inverse, a single-layer network are universal approximators when even some neurons in the network generated randomly. After that many researchers propose single-layer-based classifiers for regression and classification problem \cite{Huang2012} \cite{Zhang2012} \cite{73736251} \cite{Yang2015}.

However, the non-iterative learning algorithms for training a DCNN model are rarely found. Driven by the confliction, a more detailed motivation arises: \emph{\textbf{if the DCNN network structure maintains the same, could we use a non-iterative learning algorithm to obtain a better performance}?} Inspired by the motivation, in this paper we try to propose a non-iterative learning strategy to replace the traditional iterative learning method to boost the learning effectiveness and generalization performance further. In particular, this paper contributes the following:

1) Suitable for all DCNN models. In the proposed method, we utilize Moore-Penrose Inverse strategy to pull back the current residual error $\textbf{e}$ of the network to each fully-connected layer one by one, generating a desired output $\textbf{P}$ for each fully-connected layer. Then according to the obtained desired output and input features, we use the same strategy to recalculate weights in each fully-connected layer.  Crucially, our method only recalculates the parameters in the fully-connected layers but never involve any network structure modification, which makes the proposed method fit for all existing DCNN models.

2) Better Performance. Experimental results show that a DCNN model with the proposed method always provide better performance than the same DCNN model with its original BP method. For instance, our method achieves categorization accuracy of 94.8$\%$ on the Scene15 dataset, which is almost close to the \textbf{human-level performance}. Furthermore, as Moore-Penrose Inverse method itself does not need any iterative operation, compared to other DCNN models with iterative methods, the recomputation operation only bring a little extra computational workload (see Fig.11).

\section{The proposed method}
Training a DCNN with BP takes thousands of iterations to adjust the network parameters such as weights and biases of each layer would take several hours even in advanced GPUs. Here we show how a traditional DCNN architecture training process can be recalculated with the help of multi-layer neurons that are trained by the Moore-Penrose Inverse strategy. The detailed schematic diagram of the proposed method is shown in figure \ref{figure1}.

\subsection{DCNN with BP-based optimizer}

The convolutional layer is the core unit of modern deep learning architectures that is determined by its kernel weights that are updated during training via back-propagation. Output feature map $Cov$ w.r.t. a convolutional neural $\textbf{a}$. its associated bias $b$, and an input image/patch $\textbf{x}$ the convolutional operation is performed as
\begin{equation}
\begin{split}
Cov(m,n)=b+\sum_{k=1}^{K-1}\sum_{ip=0}^{K-1}\textbf{a}(k,i)*\textbf{x}(m+k,n+i)
\end{split}
\end{equation}
where $*$,$K$,$m,n$, and $k,i$ represent the convolutional operation, size of the kernel, first coordinate or origin of the image, and element index of the kernel respectively.

The DCNN network is trained by using Stochastic Gradient Descent with Momentum(SGDM) optimizer that minimizes binary cross-entropy loss defined by (\ref{eqn:cost_fun}), where optimizer takes a base learning rate without any particular parameter set for decaying it.

\begin{equation} \label{eqn:cost_fun}
\begin{split}
E = \frac{-1}{n} \sum\limits_{n=1}^N \left[ p_n \log \hat{p}_n + (1 - p_n) \log(1 - \hat{p}_n) \right]
\end{split}
\end{equation}

For SGDM, the gradient descent algorithm updates the parameter to minimize the error function by taking small steps in the direction of the negative gradient of the loss function.
\begin{equation}
\begin{split}
\textbf{a}_{l+1}=\textbf{a}_l+\mu\nabla E(\textbf{a}_l)
\end{split}
\end{equation}
where $l$ stands for the iteration number, $\mu$ is the learning rate, $\textbf{a}$ is the neural parameter, and $E(\textbf{a})$ is the loss function. The same as the traditional CNN methods, the gradient of the loss function, $\nabla E(\textbf{a}_l)$, is evaluated using the entire training set, and the standard gradient descent algorithm uses the entire data set at once.

\begin{figure}
  \centering
\includegraphics[width=3.5in,height=2.8in,origin=br]{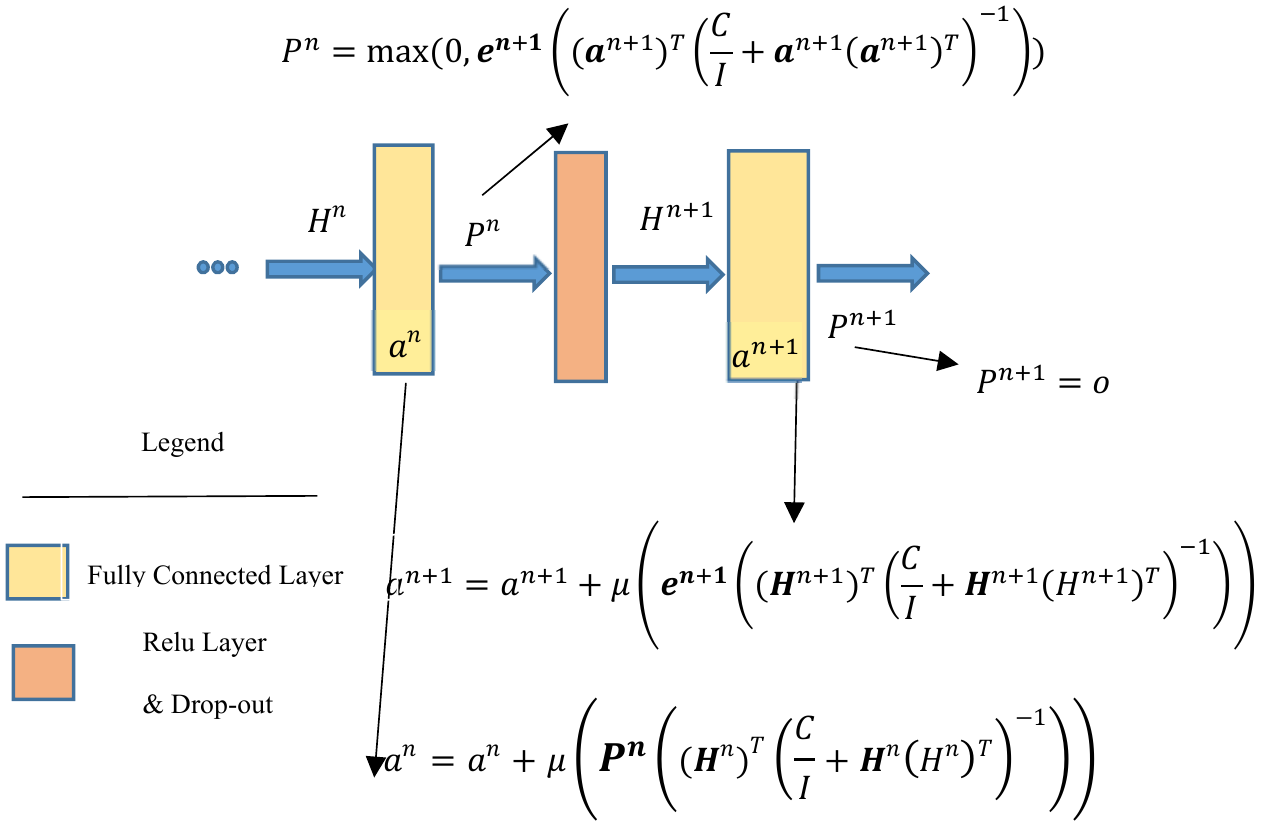}
\caption{Illustration of the key re-calculation operations with the proposed method in fully-connected layers}
\label{figure2}
\end{figure}

\begin{figure*}[!htb]
\centering
\subfigure[Details of film clips used in the experiment]{
    \label{fig:subfig:a} 
\fbox{
\begin{minipage}[b]{0.45\textwidth}
\includegraphics[width=3.2in, origin=br]{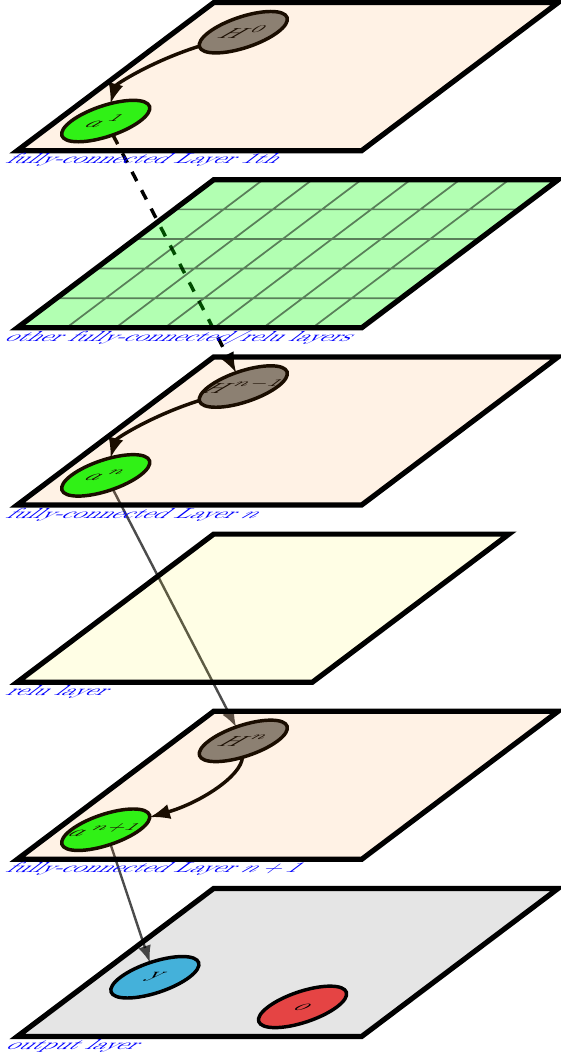}
\end{minipage}}
}
\hfill
\subfigure[the experiment scene]{
    \label{fig:subfig:b} 
    \fbox{
\begin{minipage}[b]{0.45\textwidth}
\includegraphics[width=3.2in, origin=br]{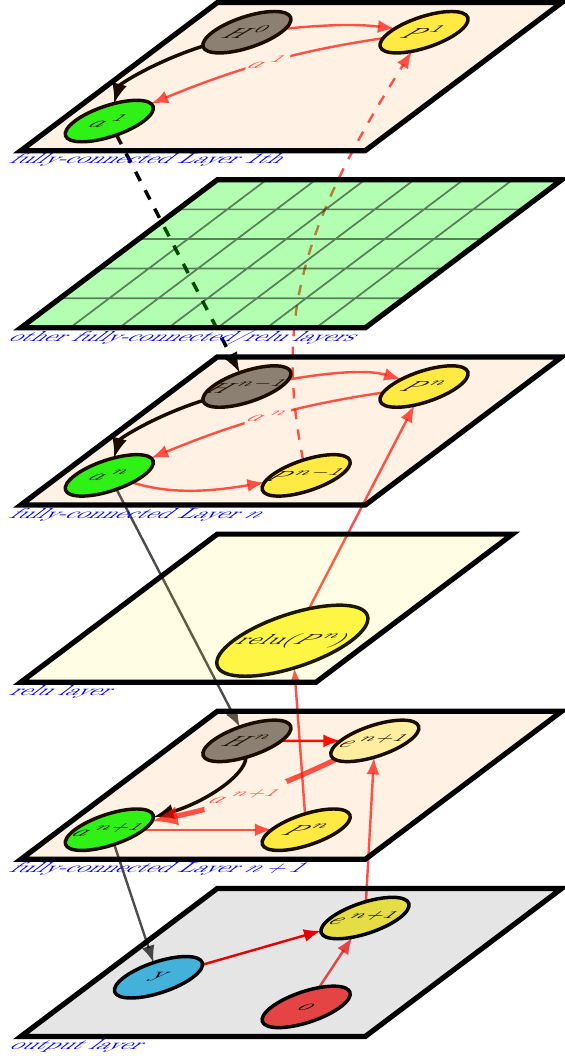}
\end{minipage}}
}
  \caption{Learning strategy of the proposed method. ${\longleftarrow}$ represents the feedforward operations of the proposed method, while $\textcolor{red}{\longleftarrow}$ represents the error-inverse operations of the proposed method. (a) Step 2-5: obtain parameters ($\textbf{a}^1,\cdots,\textbf{a}^{n+1}$)in each DCNN layer, and then extracting features ($\textbf{H}^0$) from the flatten layer, finally calculating the output of each FC layer $(\textbf{H}^1,\cdots,\textbf{H}^n$). (b) Step 6-9: recalculate parameters in each FC layer. }
  \label{learning_process}
\end{figure*}
\subsection{DCNN with the proposed method}

\subsection{Notations}

 All the notations used in the paper are shown in Table I.

\begin{table}[!htb]

\centering
\caption{Notations used in the paper}
\begin{tabular}{cccc}
\toprule
Notation & Definition     \\
\midrule
$ (\textbf{a}) $  & parameters/weights in a neuron \\
$(\textbf{a}^n$  & the parameters/weights in $n$th FC layer   \\
$\mu$ & learning rate\\
$\textbf{H}^n$ & input features of $n$th FC layer   \\
$\textbf{e}^n$ & current output error of $n$th FC layer   \\
$\textbf{P}^n$ & desired output change of $n$th FC layer   \\
$\textbf{y}$ &  output of the last FC layer   \\
$\textbf{I}$ & unit matrix\\
$\textbf{x}$ & input data\\
$\textbf{o}$ & desired output data\\
\bottomrule
\end{tabular}
\end{table}

\subsubsection{Update parameters in the soft-max layer}

Fig.\ref{figure2}-\ref{learning_process} shows our strategy to recalculate the neural parameters in FC layers. Given the desired output $\textbf{o}$, current network output ${\textbf{y}}$, the weights of the $n+1$ FC layer can be updated by Moore-Penrose inverse.

Due to
\begin{equation}
\begin{split}
\textbf{a}^{n+1}\cdot \textbf{H}^{n+1}={\textbf{y}}
\end{split}
\end{equation}
for the last FC layer ($n+1$ layer), we need to get a error-based update weight $\pmb{\eta}$, satisfying $(\textbf{a}^{n+1}+\pmb{\eta})\cdot \textbf{H}^{n+1}=\textbf{o}$. With Moore-Penrose inverse method, we can get $\pmb{\eta}$ by
\begin{equation}
\begin{split}
\pmb{\eta}&=(\textbf{o}-{\textbf{y}})\cdot(\textbf{H}^{n+1})^{-1}\\
&=(\textbf{o}-{\textbf{y}})\cdot (\textbf{H}^{n+1})^T(\frac{C}{I}+\textbf{H}^{n+1}(\textbf{H}^{n+1})^T)^{-1})\\
\end{split}
\end{equation}
As in the last FC (softmax) layer, the output error equals
\begin{equation}
\begin{split}
\textbf{e}^{n+1}=\textbf{o}-{\textbf{y}}\\
\end{split}
\end{equation}
we have
\begin{equation}
\begin{split}
\pmb{\eta}&=(\textbf{o}-{\textbf{y}})\cdot(\textbf{H}^{n+1})^{-1}\\
&=\textbf{e}^{n+1}\cdot (\textbf{H}^{n+1})^T(\frac{C}{I}+\textbf{H}^{n+1}(\textbf{H}^{n+1})^T)^{-1}\\
\end{split}
\end{equation}

Thus the last FC layer (softmax layer) can be updated by
\begin{equation}
\begin{split}
\textbf{a}^{n+1}=&\textbf{a}^{n+1}+\mu\cdot\eta\\
=&\textbf{a}^{n+1}+\mu\cdot\textbf{e}^{n+1}\cdot((\textbf{H}^{n+1})^T(\frac{C}{I}+\textbf{H}^{n+1}(\textbf{H}^{n+1})^T)^{-1}))\\
\end{split}
\end{equation}
where $\mu\in(0,1]$ represent the learning rate to overcome the over-fitting problem.

\subsubsection{Update parameters in other fully-connected layers}

As shown in Fig.2, here we already have the updated $\textbf{a}^{n+1}$ in the last FC layer, we need to obtain other values to recalculate the weights $\textbf{a}^{n}$ in $n$th FC layer. First, we try to obtain the desired output of the $n$th FC layer ($\textbf{P}^n$) throughout Moore-Penrose inverse strategy. Based on the updated $\textbf{a}^{n+1}$, we can update the current output error of $n+1$th fully-connected layer as
\begin{equation}
\begin{split}
\textbf{e}^{n+1}=\textbf{a}^{n+1}\cdot \textbf{H}^{n+1}-\textbf{o}\\
\end{split}
\end{equation}

Then we can pull the error back across the $n+1$th FC layer and the desired output change according to the updated $\textbf{e}^{n+1}$ and updated $\textbf{a}^{n+1}$ is

\begin{equation}
\textbf{P}^{n}=\textbf{e}^{n+1}\cdot(\textbf{a}^{n+1})^T(\frac{C}{I}+\textbf{a}^{n+1}(\textbf{a}^{n+1})^T)^{-1}\\
\end{equation}

Due to the relu-layer existing, we finally calculate the desired output change of the $n$th FC layer as
\begin{equation}
\begin{split}
relu(\textbf{P}^{n})&=max(0,\textbf{P}^{n})\\
\end{split}
\end{equation}

Finally the parameters in the $n$ FC layer can be recalculated by
\begin{equation}
\textbf{a}^{n}=\textbf{a}^{n}+\mu\cdot(\textbf{P}^{n}((\textbf{H}^{n})^T(\frac{C}{I}+\textbf{H}^{n}(\textbf{H}^{n})^T)^{-1}))
\end{equation}

\subsubsection{Update Parameters through a dropout operation}

Recent studies show dropout layer in DCNN also plays a vital role to counteract over-fitting issue. With a dropout operation, the parameters in each fully-connected layer could be updated by
\begin{equation}
\textbf{a}^{n}=\digamma(\textbf{a}^{n}+\mu\cdot(\textbf{P}^{n}((\textbf{H}^{n})^T(\frac{C}{I}+\textbf{H}^{n}(\textbf{H}^{n})^T)^{-1})))
\end{equation}
where $\digamma$ represent a dropout operation to partially update neurons with a random selection way. For example, if we randomly choose a $50\%$ dropout rate, the detailed operation steps can be indicated as the following Fig.4.

\begin{figure}[!htb]
  \centering
\includegraphics[width=3.5in,height=3in, origin=br]{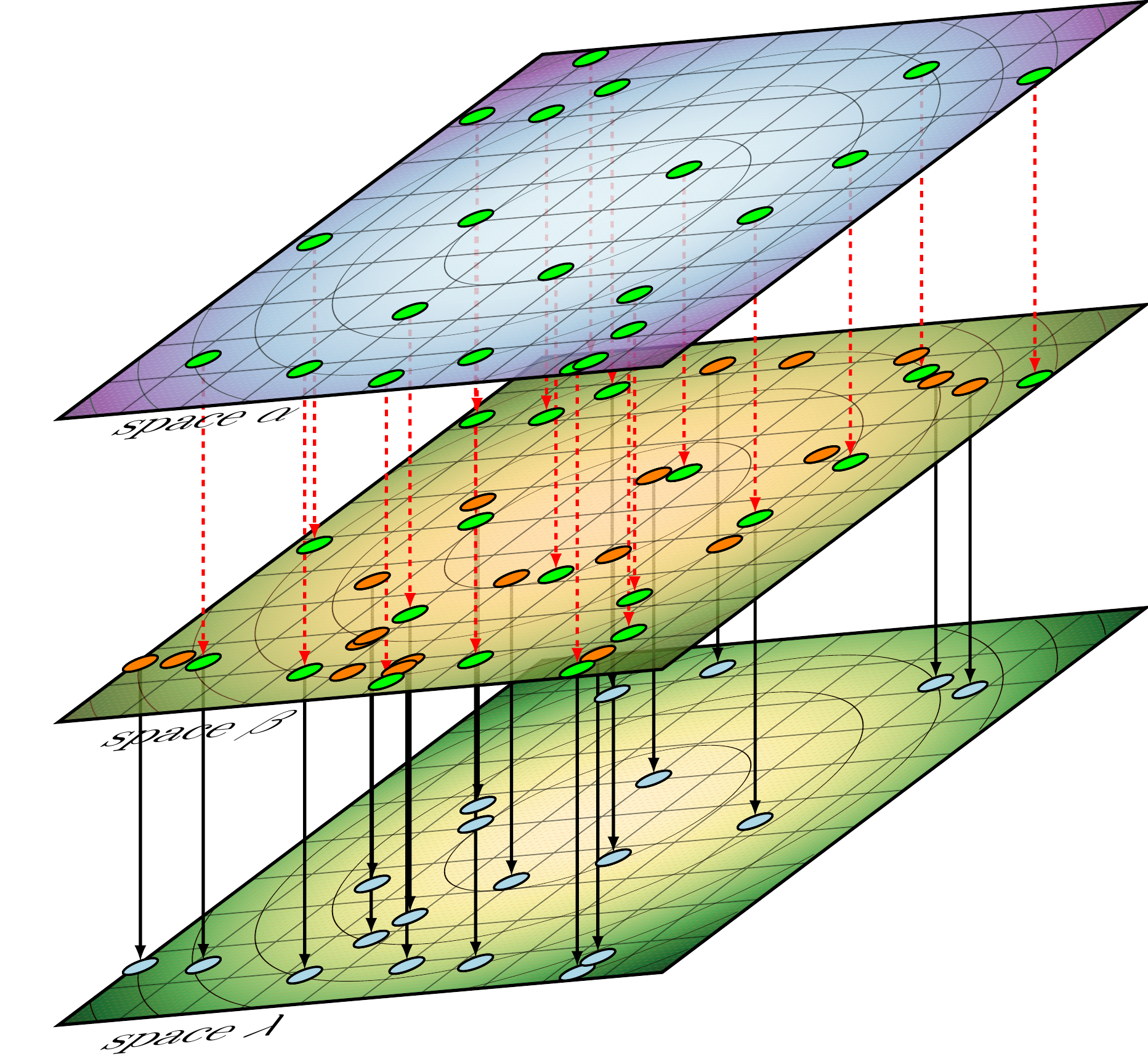}
  \caption{Dropout strategy for updating parameters.  For instance, the initial parameters of 20 neurons are distributed in space $\alpha$. After updating the neurons by the proposed method, the parameters of 20 neurons are distributed in space $\beta$. Finally, if the dropout rate equals 50 percent, we only randomly select ten neurons from space $\alpha$, and the other ten neurons from space $\beta$ to finalize the 20 recalculated neurons.}
\end{figure}

\subsubsection{The learning steps of the proposed method}

Based on the equations (8)-(13), our method can be summarized as following.

Step 1: Given a designed DCNN network architecture, input images dataset with labels $\textbf{x},\textbf{o}$, positive coefficient $C$, learning rate $\mu$, momentum $m$, and maximum training epoch number $L$.

Step 2: Use SGDM-optimizer to train the DCNN network with only one training epoch.

Step 3: Extract deep features from the flatten layer ($\textbf{H}^0$).

Step 4: Obtain the current output of each FC layer ($\textbf{H}^1,\cdots,\textbf{H}^{n+1}$) by
\begin{equation}
\begin{split}
&\textbf{H}^n={\textbf{a}^n}\cdot\textbf{H}^{n-1}\\
\end{split}
\end{equation}

Step 5: Obtain the current output error ($\textbf{e}^{n+1}$) of the $n+1$th FC layer by equation (6)

Step 6: recalculate the parameter ($\textbf{a}^{n+1}$) of the $n+1$th FC layer via equation (8).

Step 7: Obtain the desired output change ($\textbf{P}^n$) of the $n$th FC layer by equation (9)-(10).

Step 8: recalculate the parameter ($\textbf{a}^n$) of the $n$th FC layer by equation (13)

Step 9: recalculate parameters ($\textbf{a}^1,\cdots, \textbf{a}^{n-1}$) in other FC layers through Step 7-9.

Step 10: Use Step 2-9 $L-1$ times to obtain the finalize the trained DCNN model.

\begin{table*}[!htb]
 \newsavebox{\tablebox}
\centering
\caption{Specification of used image datasets in the experiment}
\begin{lrbox}{\tablebox}
\begin{tabular}{cccccc}
\toprule
Datasets      &\#\makecell{Training images\\ per Category}  &\#total training image &\#total testing/validation image &\# Category  \\
\midrule
Scene15   &100 &1,500 &2,985 &15   \\
SUN397   &50 &19,850 &19,850 &397   \\
Cifar10  &5,000 &50,000 &10,000 &10   \\
Cifar100   &500 &50,000 &10,000 &100   \\
Places365(500 images per class)   &500 &182,500 &36,500 &365   \\
Places365(1000 images per class)   &1,000 &365,000 &36,500 &365   \\
Places365(1500 images per class)  &1,500 &547,500 &36,500 &365   \\
ImageNet Mini   &200 &200,000 &50,000 &1000\\
ImageNet    &7,32-1,300 &1,281,168 &50,000 &1000\\
\bottomrule
\end{tabular}
\label{table1}
\end{lrbox}
\scalebox{0.9}{\usebox{\tablebox}}
\end{table*}

\begin{table}[!htb]
\caption{Scene-15 classification accuracy for our method against leading alternate approaches without data argumentation}
\begin{lrbox}{\tablebox}
\begin{tabular}{ll}
\toprule
Method & Scene15\\
\midrule
 \emph{Improved classifiers based on NN/SVM/Kernel/KNN} & \\
\,\,\,\,\,\,\,\,\,\,Kernel codebook \cite{GemertECCV08} & 76.6\\
\,\,\,\,\,\,\,\,\,\,Object-to-class kernels \cite{ZhangL2014} &88.8\\
\,\,\,\,\,\,\,\,\,\,KNN with localized multiple kernel \cite{Han2014} & 89.1\\
\,\,\,\,\,\,\,\,\,\,Label Consistent K-SVD, Spatial pyramid \cite{JiangPAMI13} & 92.9\\
\midrule
 \emph{Sparse representation-based methods} & \\
\,\,\,\,\,\,\,\,\,\,Linear spatial pyramid, sparse coding \cite{YangCVPR2009} &80.3   \\
\,\,\,\,\,\,\,\,\,\,Laplacian sparse coding, feature combination \cite{GaoCVPR2010} & 88.9\\
\midrule
\emph{Recent feature coding methods} & \\
\,\,\,\,\,\,\,\,\,\,Feature fusion \cite{YuPR13} & 71.6\\
\,\,\,\,\,\,\,\,\,\,Visual word ambiguity \cite{VanPAMI08} & 76.7\\
\,\,\,\,\,\,\,\,\,\,Hard assignment \cite{LazebnikCVPR06} &81.4  \\
\,\,\,\,\,\,\,\,\,\,Soft assignment \cite{LingqiaoICCV11} &82.2\\
\,\,\,\,\,\,\,\,\,\,Centrist, Spatial PACT \cite{WuRehg2011} &83.9\\
\midrule
\emph{Hierarchical networks} & \\
\,\,\,\,\,\,\,\,\,\,Feature pooling \cite{BoureauICML10} &80.6 \\
\,\,\,\,\,\,\,\,\,\,Multilayer ELM, SIFT features \cite{Yang2015} &82.4\\
\,\,\,\,\,\,\,\,\,\,Sparse coding, Max-pooling\cite{Boureau2010} &84.3\\
\,\,\,\,\,\,\,\,\,\,Six-layer deep network, Macro Feature  \cite{Goh2014} &85.4\\
\,\,\,\,\,\,\,\,\,\,Five-layer manifold deep network \cite{yuanyuan2015} &86.9\\
\midrule
\emph{CNN networks with pre-trained features} & \\
\,\,\,\,\,\,\,\,\,\,Hybrid-CNN, pretrained by Places205 dataset \cite{Zhounips2014} &91.6\\
\,\,\,\,\,\,\,\,\,\,AlexNet, pretrained by Places365 dataset \cite{Zhoubolei16} &90.0\\
\,\,\,\,\,\,\,\,\,\,AlexNet, pretrained by ImageNet dataset  &82.4\\
\,\,\,\,\,\,\,\,\,\,GoogLeNet, pretrained by Places365 dataset \cite{Zhoubolei16} &91.2\\
\,\,\,\,\,\,\,\,\,\,16-layer VGG, pretrained by Places365 dataset \cite{Zhoubolei16} &92.0\\
\,\,\,\,\,\,\,\,\,\,16-layer VGG, pretrained by ImageNet dataset &88.0\\
\midrule
\emph{Our architecture} & \\
\,\,\,\,\,\,\,\,\,\,Our method with ImageNet pretrained Alexnet  & \textbf{86.2} \\
\,\,\,\,\,\,\,\,\,\,Our method with ImageNet pretrained 16-layer VGG  & \textbf{89.8} \\
\,\,\,\,\,\,\,\,\,\,Our method with Places205 pretrained Alexnet  & \textbf{91.8} \\
\,\,\,\,\,\,\,\,\,\,Our method with Places205 pretrained 16-layer VGG  & \textcolor{blue}{\textbf{94.8}} \\
\midrule
\,\,\,\,\,\,\,\,\,\,\textbf{Human-level Performance}\tablefootnote{\cite{Zhounips2014} mentioned "This dataset (Scene15) contains only 15 scene categories with a few hundreds images per class, where current classifiers are saturating this dataset nearing human performance at 95 percent".}   \cite{Zhounips2014}  & \textcolor{blue}{\textbf{95.0}} \\
\bottomrule
\end{tabular}
\end{lrbox}
\scalebox{0.99}{\usebox{\tablebox}}
\label{Table6}
\end{table}

\begin{figure}[!htb]
  \begin{center}
  \begin{minipage}{\columnwidth}
\begin{tikzpicture}[scale=0.38]
legend style={text=black, font=\fontsize{7}{8}\selectfont},

\begin{drawaxes}[
diagram angle=15,
diagram radius=10,
draw axes]{};
\end{drawaxes}

\sectorlist[sector angle=30,
draw sector list={3cm/red!90!blue,  
  3.4cm/magenta,   
  3.8cm/magenta!40!violet,   
  4.2cm/violet!50!blue,   
  5.4cm/blue!80!cyan,   
  5.8cm/blue!20!cyan,   
  6.7cm/cyan!50!green,  
  5.9cm/green!85!blue,  
  5.7cm/green!90!lime,  
  5.35cm/yellow,   
  4.8cm/orange,   
  4.1cm/red!70!orange  
  }]{};

\begin{customlegend}[
legend entries={
Ours with VGG16,Densely Net\cite{Huang_2017_CVPR},VGG16,Inception Net,All Convnet\cite{ICLRJost},160-Resnet\cite{Liang2015},
AlexNet\cite{Zhoubolei16},Inception Net,Hybrid-CNN\cite{Zhounips2014},VGG16,Label K-SVD\cite{JiangPAMI13},Ours with VGG16
},
legend style={at={(-13.5,-8)},anchor=south west,font=\scriptsize}]
\legendlistcolors{red!90!blue,magenta,magenta!40!violet,violet!50!blue,blue!80!cyan,blue!20!cyan,cyan!50!green,green!85!blue,green!90!lime,yellow,orange,red!70!orange}
\end{customlegend}
\end{tikzpicture}

\end{minipage}

  \end{center}
  \caption{Comparison Testing Error Rate on CIFAR10 and Scene15 dataset: Our method vs other leading CNN models. Sectors in positive $y$ axis represent error rate on Scene15; Sectors in negative $y$ axis represent error rate on CIFAR10.}
  \label{fig:fom}
\end{figure}
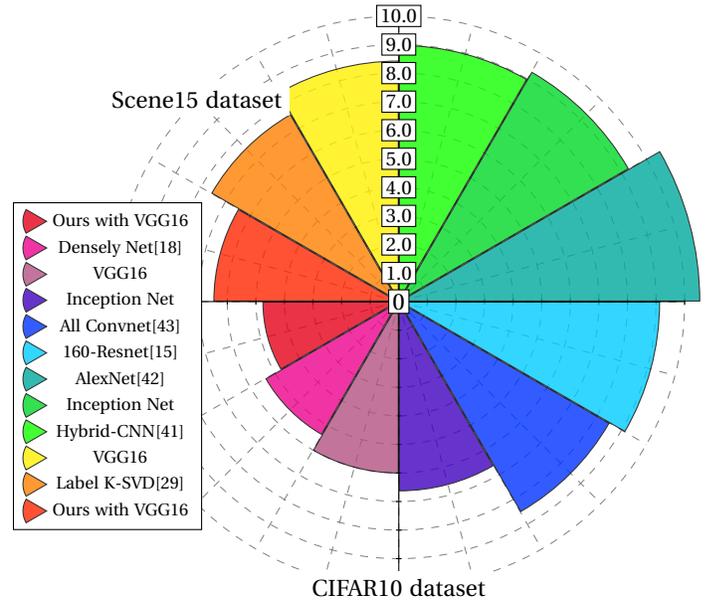

\begin{figure}[!htb]
  \centering
\includegraphics[width=3.5in,height=3in, origin=br]{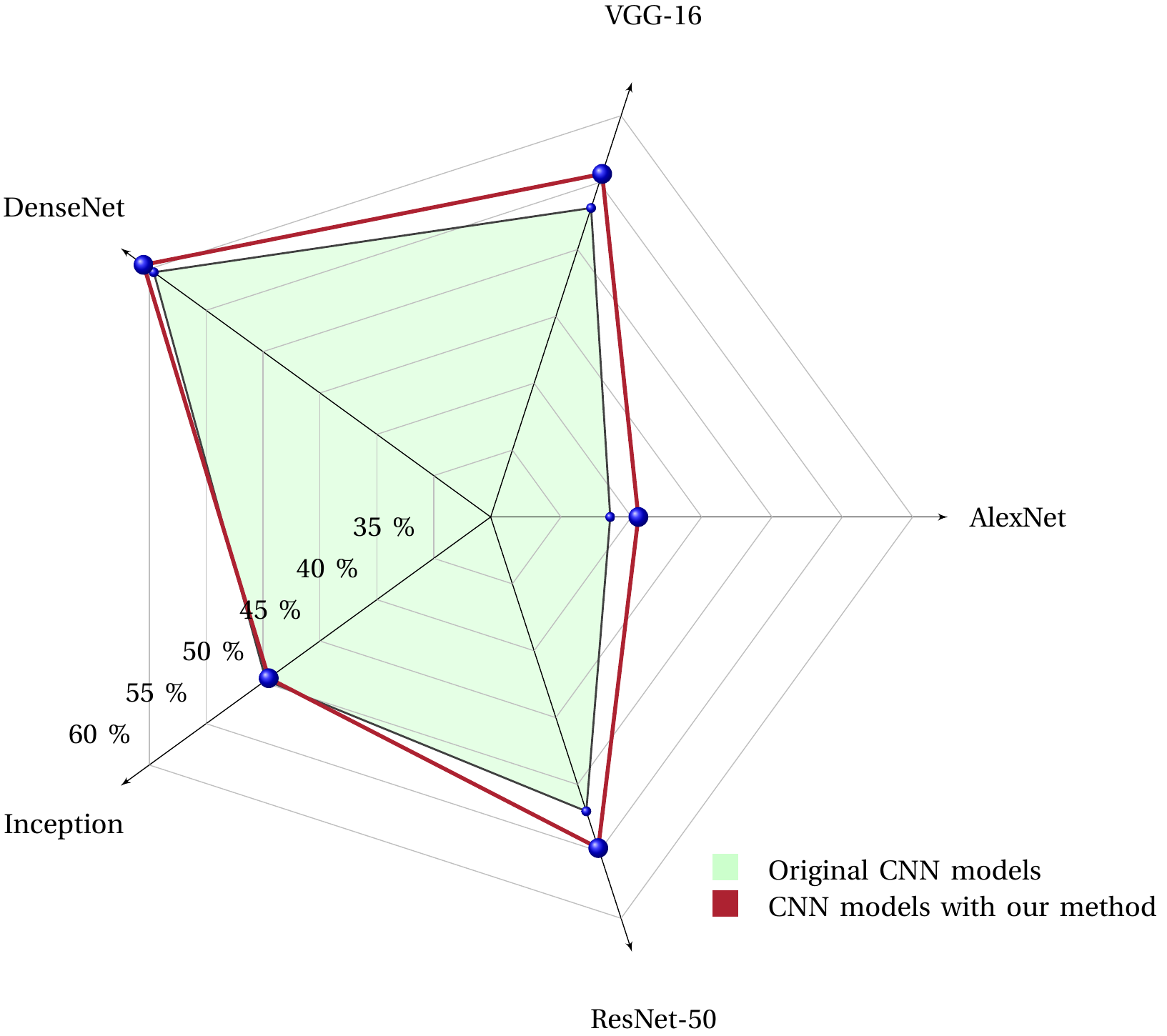}
  \caption{Performance difference on Sun397 dataset.}
\end{figure}

    \begin{figure*}
        \centering
        \begin{tikzpicture}
            \begin{groupplot}[
            legend style={text=black, font=\fontsize{7}{8}\selectfont},
                    grid=both,
                    ybar,
                    legend to name=CombinedLegendBar,
                    ylabel={Testing error rate ($\%$)},
                    xlabel={},
                    view={0}{90},
                    width=14cm,
                    height=7cm,
                    symbolic x coords={1998,2002,2006,2007,2009,2010,2012},
                    xtick=data,
                    every node near coord/.append style={font=\large},
                    nodes near coords,
                    nodes near coords align={vertical},
                    group style={
                    group size=2 by 1,
                    xlabels at=edge bottom,
                    ylabels at=edge left,
                    xticklabels at=edge bottom}]

                \nextgroupplot[]
                   \addplot[ybar,pattern=horizontal lines, pattern color=blue] coordinates {(1998, 1.1) (2002, 0.68) (2006, 0.60)(2007, 0.56) (2009, 0.53) (2010, 0.35)(2012, 0.23) };

    \addlegendentry{AlexNet}
    \addlegendentry{16-layer VGG Model}
    \addlegendentry{GoogleNet}
    \addlegendentry{50-layer ResNet}
    \addlegendentry{Our model}

            \end{groupplot}

        \end{tikzpicture}
        \ref{CombinedLegendBar}
        \caption{Testing Accuracy Performance Comparison between our methods with Inception model and Original Inception model.}
        \label{PlusPlusCombinedBar}
    \end{figure*}
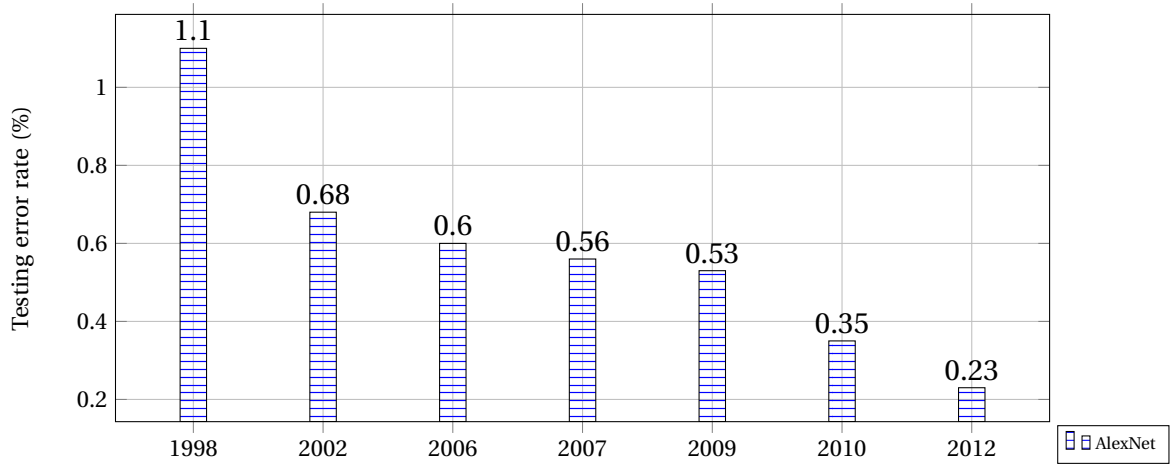

\begin{figure}[!htb]
  \centering
\includegraphics[width=3.5in,height=3in, origin=br]{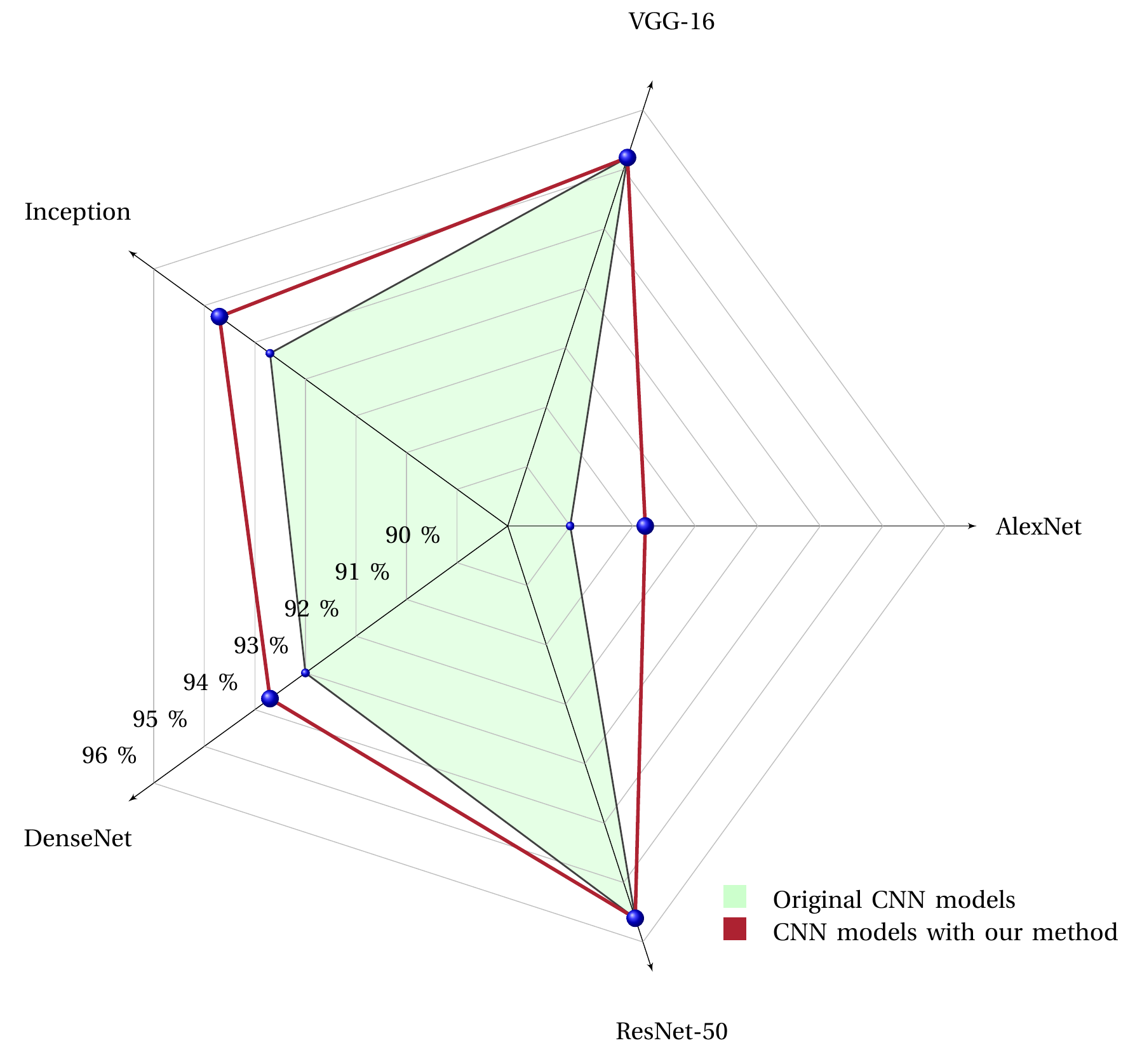}
  \caption{Performance difference on CIFAR10 dataset.}
\end{figure}

\begin{figure}[!htb]
  \centering
\includegraphics[width=3.5in,height=3in, origin=br]{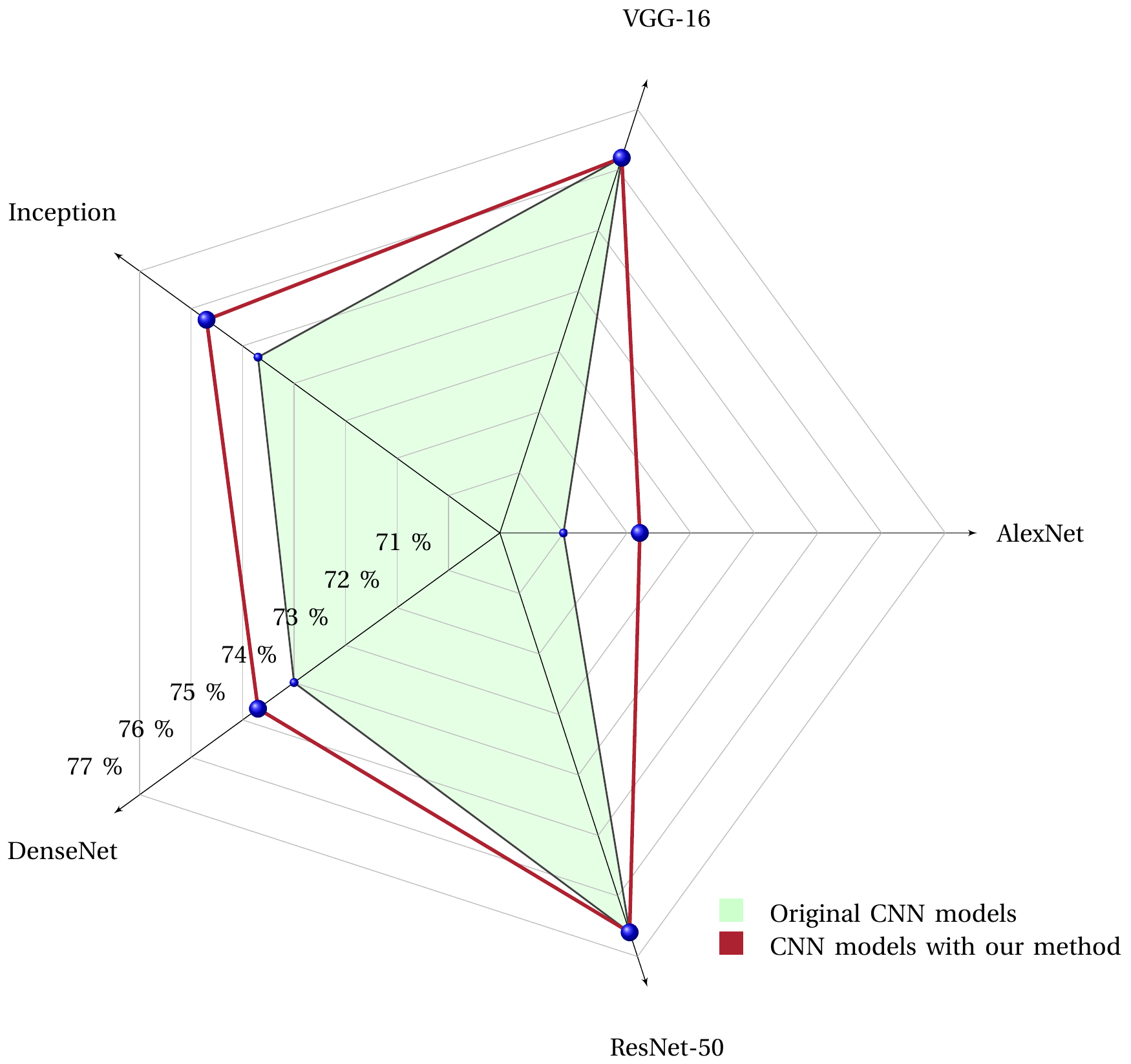}
  \caption{Performance difference on CIFAR100 dataset.}
\end{figure}

\section{Experimental Verification}

\subsection{Rival methods and experiment environment settings}

In this section aimed at examining the performance of our proposed learning method, we test the proposed method on several image datasets. The experiments are conducted in two environments (Matlab 2017b or Keras). For the complete comparisons, we evaluate the \textbf{29} art-of-the state methods arising from the following \textbf{three families}:

(1) Recent feature coding methods with single layer classifiers include hard/soft assignment \cite{LazebnikCVPR06}, centrist \cite{WuRehg2011}, feature fusion \cite{YuPR13}, sparse coding \cite{YangCVPR2009},  Laplacian sparse coding \cite{GaoCVPR2010}, kernel codebook \cite{GemertECCV08}, Object-to-class kernels \cite{ZhangL2014}, multilayer ELM \cite{Yang2015}, K-SVD \cite{JiangPAMI13}, Visual word ambiguity \cite{VanPAMI08}, Soft assignment \cite{LingqiaoICCV11}, and KNN Kernel \cite{Han2014}.

(2) Recent DCNN models include AlexNet \cite{NIPS7433451}, VGG-16/19 model \cite{Simonyan14c}, 96/160-layer recurrent convolutional network \cite{Liang2015}, Network in Networks \cite{NINyan}, Google-Inception model \cite{7298594}, ResNet, Densely CNN model \cite{Huang_2017_CVPR}, ImageNet-pretrained CNN models, Places-pretrained models \cite{Zhoubolei16}, hybrid-CNN features \cite{Zhounips2014}, hierarchical manifold deep network \cite{yuanyuan2015}, Multi-column deep network \cite{Schmidhuber:2012:MDN:2354409.2354694}, All convolutional net \cite{ICLRJost},  and Deep-supervised Nets \cite{LeeNiN}.

(3) Recent DCNN with feature fusion technologies include multi-layer deep network \cite{Goh2014}, max-pooling with spatial pyramid features \cite{Boureau2010}, feature pooling \cite{BoureauICML10}, Deep attention selective networks \cite{NIPS2014_5276}, sumproduct network with deep architecture \cite{NIPS2012_4516}.

\subsection{Experimental Environment and Datasets Selection}

In the experiment, we select some widely used image datasets to evaluate our method. For completeness, we select six image databases which show in Table 1, including one small dataset (Scene15), three medium datasets (CIFAR10/100, SUN397), and two large datasets (Places365 and ImageNet). For small/medium dataset tests, the experiments are conducted in Matlab 2017b or Keras with 32 GB of memory, Geforce 1080 8GB GPU, and an I7-4470 (3.4G) processor. For the large-scale datasets, a workstation with 128 GB memory, and one Geforce 1080 TI GPU is used to run the test. All the results are obtained over three trials. To highlight general trends, we mark all results that outperform the existing state-of-the-art in boldface and the best result in blue color.

\textbf{Scene15} dataset contains 4486 gray-value images, of which 3860 images are from the 13-category scenes in \cite{Fei-fei}. Each category has 200 to 400 images, and the average image size is about $250\times 300$ pixels. Following the common experimental settings, we randomly select 100 images per category as training data and use the rest as test data.

\textbf{CIFAR10/100}. The two CIFAR datasets consist of natural colored images with $32\times 32$ pixel. Cifar10 consists of 60,000 color images in 10 categories including airplane, bird, automobile, cat, dog, frog, deer, ship, horse, and truck. Cifar100 consists 60,000 color images in 100 classes. There are 50,000 images (5,000 per class) for training and the remaining parts for testing. For the average results, we use all 50,000 training images and report the final testing accuracy without any data argumentation strategy.

\textbf{SUN397} is a medium scene recognition dataset with about 100k images spanning 397 categories. According to some previous researchers \cite{7805494}  \cite{DBLP:journals/corr/ZhengZWWT16}, we randomly split the datasets into a training dataset and a testing dataset, each with 50 images per class. Thus there are 19,850 images for both the training and testing dataset.

\textbf{Places365} \cite{zhou2017places} is a large-scale dataset extended from Places205. In total, Places365 contains around 1,800,000 images comprising 365 unique scene categories. The dataset features at least 3000 training images per class, consistent with real-world frequencies of occurrence. To further test our method with other CNN models, comparison experiments are conducted on the Places365. Due to the computing resource limitation, we randomly select 500,1000, and 1500 images per class from the original Places365 training set to produce the training set. We also use the validation dataset within 36,500 images to generate a testing dataset. The detailed information about the database is shown in Table 1.

\textbf{ImageNet} \cite{ILSVRC15} is a large-scale dataset, which consists around 1.2 million images for training, and 50,000 for validation, from 1,000 classes. We adopt the same data augmentation scheme for training images as in \cite{Huang_2017_CVPR}, and apply a single-crop with size 224$\times$224 at test time. Due to the training time limitation, at the current stage, we only carry out the comparative performance between BP-based DenseNet and DenseNet with our method. Due to required huge training time (17 days per time with single GPU), We will add more relative comparative results based on other well-known DCNN models including AlexNet, VGG, and ResNet in the final publication.

    \begin{figure*}[!htb]
        \centering
        \begin{tikzpicture}
            \begin{groupplot}[
            legend style={text=black, font=\fontsize{7}{8}\selectfont},
            grid=both,
                    legend columns=-1,
                    legend entries={{\color{black}{AlexNet Model}},{\color{black}{Our method with AlexNet Model}}, LR represents Learning Rate.},
                    legend to name=CombinedLegendBar,
                    ylabel={Testing Accuracy on Places365},
                    xlabel={\# Training Epoch},
                    width=6.5cm,
                    height=7cm,
                    ybar legend,
                    symbolic x coords={1,2,3,4,5,6,7},
                    xtick=data,
                    every node near coord/.append style={font=\tiny},
                    nodes near coords,
                    nodes near coords align={vertical},
                    group style={
                    group size=3 by 1,
                    xlabels at=edge bottom,
                    ylabels at=edge left,
                    xticklabels at=edge bottom}]

                \nextgroupplot[title={500 training samples per class}]
                   \addplot[ybar, color=blue, fill=yellow!80!white!45, ybar legend] coordinates {(1, 25.38) (2, 27.24) (3, 35.47) (4, 35.63) (5, 36.15) (6,36.28) };
                    \addplot [sharp plot, color=brown, thick,
                 mark=*, mark options={fill=blue, scale=1},text mark as node=true,
                 thick, mark size=2.0pt] coordinates {(1,25.58) (2, 27.19) (3,  36.40) (4, 36.68) (5,38.25) (6, 38.71)};
                 \draw [<->] [auto, thick,rounded corners=5pt, color=black] (-50,30) -- (80,30)node[above]{\tiny learning rate $=1.0^{-3}$} -- (100,30) ;
                 \draw [<->] [auto, thick,rounded corners=5pt, color=black] (100,50) -- (200,50)node[above]{\tiny learning rate $=1.0^{-4}$} -- (300,50) ;
                 \draw [<->] [auto, thick,rounded corners=5pt, color=black] (300,60) -- (400,60)node[above]{\tiny learning rate $=1.0^{-5}$} -- (500,60) ;

                \nextgroupplot[title={1000 training samples per class}]
                    \addplot[ybar, color=blue, fill=yellow!80!white!45, ybar legend] coordinates {(1, 27.74) (2, 30.21)  (3, 38.00) (4, 38.50) (5, 39.10)(6, 39.16) };
                    \addplot [sharp plot, color=brown, thick,
                 mark=*, mark options={fill=blue, scale=1},text mark as node=true,
                 thick, mark size=2.0pt] coordinates {(1,31.62) (2, 31.67) (3,  39.61) (4,  40.56) (5, 40.90) (6,40.92)};
                 \draw [<->] [auto, thick,rounded corners=5pt, color=black] (-50,30) -- (80,30)node[above]{\tiny learning rate $=1.0^{-3}$} -- (100,30) ;
                 \draw [<->] [auto, thick,rounded corners=5pt, color=black] (100,50) -- (200,50)node[above]{\tiny learning rate $=1.0^{-4}$} -- (300,50) ;
                 \draw [<->] [auto, thick,rounded corners=5pt, color=black] (300,60) -- (400,60)node[above]{\tiny learning rate $=1.0^{-5}$} -- (500,60) ;
                \nextgroupplot[title={1500 training samples per class}]
                    \addplot[ybar, color=blue, fill=yellow!80!white!45, ybar legend] coordinates {(1, 30.32) (2, 29.81) (3, 38.79) (4, 39.45) (5, 40.05) (6, 40.13)};
                    \addplot [sharp plot, color=brown, thick,
                 mark=*, mark options={fill=blue, scale=1},text mark as node=true,
                 thick, mark size=2.0pt] coordinates {(1,31.22) (2, 32.07) (3, 40.49 ) (4,  41.50) (5, 42.21) (6,42.21)};
                 \draw [<->] [auto, thick,rounded corners=5pt, color=black] (-50,30) -- (80,30)node[above]{\tiny learning rate $=1.0^{-3}$} -- (100,30) ;
                 \draw [<->] [auto, thick,rounded corners=5pt, color=black] (100,50) -- (200,50)node[above]{\tiny learning rate $=1.0^{-4}$} -- (300,50) ;
                 \draw [<->] [auto, thick,rounded corners=5pt, color=black] (300,60) -- (400,60)node[above]{\tiny learning rate $=1.0^{-5}$} -- (500,60) ;

            \end{groupplot}
        \end{tikzpicture}
        \ref{CombinedLegendBar}
        \caption{ Top-1 Testing Accuracy of Places365: Our method with AlexNet vs AlexNet. }
        \label{PlusPlusCombinedBar}
    \end{figure*}
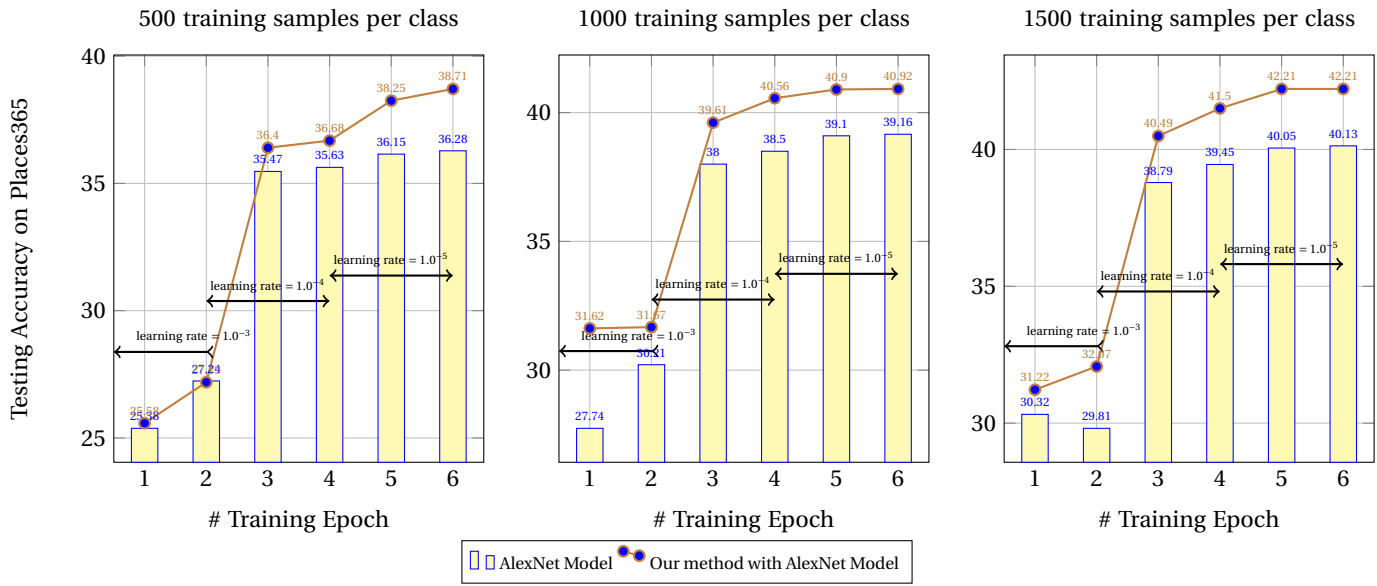

We compare our method with other state-of-the-art methods in two ways. (i) Our method vs. other classifiers; (ii) the DCNN model with our method vs. the same original DCNN model. For the results category (i) (in subsection 3.3), we try to indicate that the proposed method could, in general, provide very comparable results among the recent well-known image recognition methods. More importantly, for the results category (ii) (in subsection 3.4), we are going to demonstrate that any DCNN network with our proposed method could, in general, bring a better generalization performance than that same model with BP-based methods.

\subsection{Comparison performance of DCNN with our method vs. other 31 state-of-the-art methods}

\begin{table*}[!htb]
\small
\centering
\caption{Classification accuracy for our method against other leading methods without data augmentation}
\begin{lrbox}{\tablebox}
\begin{tabular}{cccc}
\toprule
Method & CIFAR10 & CIFAR100\\
\midrule
         \emph{Hierarchical networks}  \\

\,\,\,\,\,\,\,\,\,\,sumproduct network with deep architecture \cite{NIPS2012_4516} &84.1  &-  \\
\,\,\,\,\,\,\,\,\,\,Multi-column deep networks \cite{Schmidhuber:2012:MDN:2354409.2354694} &88.9 &- \\
\,\,\,\,\,\,\,\,\,\,Deep attention selective networks \cite{NIPS2014_5276} &90.7 &66.3 \\
\,\,\,\,\,\,\,\,\,\,Deep-supervised Nets \cite{LeeNiN} &90.2 &- \\
\,\,\,\,\,\,\,\,\,\,96-layers Recurrent convolutional network \cite{Liang2015} &89.7 &65.8 \\
\,\,\,\,\,\,\,\,\,\,160-layers Recurrent convolutional network \cite{Liang2015} &91.3 &68.3 \\
\,\,\,\,\,\,\,\,\,\,44-layers ResNet \cite{He_2016_CVPR} &92.8 &- \\
\,\,\,\,\,\,\,\,\,\,110-layers ResNet \cite{He_2016_CVPR} &93.5 &- \\
\,\,\,\,\,\,\,\,\,\,Network in Networks, pretrained by ImageNet dataset \cite{NINyan} &89.6 & 64.4  \\
\,\,\,\,\,\,\,\,\,\, Densely connected convolutional networks \cite{Huang_2017_CVPR} & 94.8  &{80.4}  \\
 \,\,\,\,\,\,\,\,\,\,All convolutional net, ImageNet-pretrained \cite{ICLRJost} &92.0  &75.6 \\

      \midrule
\emph{Our method with CNN models} & \\
\,\,\,\,\,\,\,\,\,\,Ours with AlexNet &{91.2}  &{74.5}  \\
\,\,\,\,\,\,\,\,\,\,Ours with Google Inception &{{94.7}}  &{77.3} \\
\,\,\,\,\,\,\,\,\,\,Ours with 16-layer VGG &{{95.2}}  &{80.1} \\
\,\,\,\,\,\,\,\,\,\,Ours with ResNet-50 &{\textbf{95.6}}  &{\textbf{81.9}} \\
\bottomrule
\end{tabular}
\end{lrbox}
\scalebox{0.9}{\usebox{\tablebox}}
\label{Table3_new}
\end{table*}

\begin{table*}[!htb]
\small
\centering
\caption{Classification accuracy for CNN models with our method against BP based CNN models}
\begin{lrbox}{\tablebox}
\begin{threeparttable}[h]
\begin{tabular}{cccc}
\toprule
Method & CIFAR10 & CIFAR100 &SUN397\\
\midrule
         \emph{Pretrained AlexNet}  \\
\ \,\,\,\,\,\,\,\,\,\,AlexNet \cite{NIPS7433451}, ImageNet-pretrained  &90.0  &73.1 &38.5\\
\,\,\,\,\,\,\,\,\,\,Ours with ImageNet-pretrained AlexNet &\textbf{91.2}  &\textbf{74.5} &\textbf{40.5} \\
\ \,\,\,\,\,\,\,\,\,\,AlexNet, Places365-pretrained  &-  &- &39.1\\
\,\,\,\,\,\,\,\,\,\,Ours with Places365-pretrained AlexNet &-  &- &\textbf{42.0} \\
      \midrule
                  \emph{Pretrained VGG-16}  \\
\,\,\,\,\,\,\,\,\,\,16-VGG, ImageNet-pretrained \cite{Simonyan14c}  &95.2  &79.4 &53.1\\
\,\,\,\,\,\,\,\,\,\,Ours with ImageNet-pretrained 16-layer VGG &95.2  &\textbf{80.1} & {\textbf{55.6}}\\
      \midrule
                  \emph{Google Inception}  \\
\,\,\,\,\,\,\,\,\,\,Google-Inception, ImageNet-pretrained \cite{7298594}  &93.7  &77.1 &\textbf{49.3}\\
 \,\,\,\,\,\,\,\,\,\,Ours with ImageNet-pretrained Google Inception &{\textbf{94.7}}  &\textbf{77.3} & {48.9}\\
       \midrule
                   \emph{DenseNet}  \\
\,\,\,\,\,\,\,\,\,\, 40-layer DenseNet, training from scratch \cite{Huang_2017_CVPR} & 93.0\cite{Huang_2017_CVPR}  &72.6\cite{Huang_2017_CVPR} &59.6$^a$ \\
 \,\,\,\,\,\,\,\,\,\,Ours with 40-layer DenseNet,training from scratch &{\textbf{93.7}}  &\textbf{73.3} & {\textbf{60.4}$^a$ }\\
      \midrule
                   \emph{ResNet-50}  \\
    \,\,\,\,\,\,\,\,\,\,ImageNet-pretrained ResNet-50 &95.2  &80.8 &52.0\\
\,\,\,\,\,\,\,\,\,\,Ours with ImageNet-pretrained ResNet-50 &\textbf{95.6}  &\textbf{81.9} & \textbf{54.3}\\
\bottomrule
\end{tabular}
\begin{tablenotes}
\item [a] 121-layer ImageNet Pretrained DenseNet .
\end{tablenotes}
\end{threeparttable}
\end{lrbox}
\scalebox{0.9}{\usebox{\tablebox}}
\label{Table3_new}
\end{table*}

    \begin{figure*}[!htb]
        \centering
        \begin{tikzpicture}
            \begin{groupplot}[
            legend style={text=black, font=\fontsize{7}{8}\selectfont},
            grid=both,
                    legend columns=-1,
                    legend entries={{\color{black}{VGG-16 Model}},{\color{black}{Our method with VGG-16 Model}}},
                    legend to name=CombinedLegendBar,
                    ylabel={Testing Accuracy on Places365},
                    xlabel={\# Training Epoch},
                    view={0}{90},
                    width=6.5cm,
                    height=7cm,
                    ybar legend,
                    symbolic x coords={1,2,3,4,5,6,7},
                    xtick=data,
                    every node near coord/.append style={font=\tiny},
                    nodes near coords,
                    nodes near coords align={vertical},
                    group style={
                    group size=3 by 1,
                    xlabels at=edge bottom,
                    ylabels at=edge left,
                    xticklabels at=edge bottom}]

                \nextgroupplot[title={500 training samples per class}]
                   \addplot[ybar, color=brown, fill=yellow!80!white!45, ybar legend] coordinates {(1, 36.85) (2, 40.78 ) (3, 45.15) (4, 44.62) (5, 44.73) (6,44.69) };
                    \addplot [sharp plot, color=blue, thick,
                 mark=*, mark options={fill=red, scale=1},text mark as node=true,
                 thick, mark size=2.0pt] coordinates {(1,38.67 ) (2, 42.51) (3, 46.14) (4, 46.26  ) (5, 46.01) (6,46.05)};
                 \draw [<->] [auto, thick,rounded corners=5pt, color=black] (-50,50) -- (80,50)node[above]{\tiny learning rate $=1.0^{-3}$} -- (100,50) ;
                 \draw [<->] [auto, thick,rounded corners=5pt, color=black] (100,150) -- (200,150)node[above]{\tiny learning rate $=1.0^{-4}$} -- (300,150) ;
                 \draw [<->] [auto, thick,rounded corners=5pt, color=black] (300,300) -- (400,300)node[above]{\tiny learning rate $=1.0^{-5}$} -- (500,300) ;p

                \nextgroupplot[title={1000 training samples per class}]
                    \addplot[ybar, color=brown, fill=yellow!80!white!45, ybar legend] coordinates {(1, 40.85) (2, 43.59) (3, 47.41) (4, 47.41) (5, 47.43)(6, 47.34) };
                    \addplot [sharp plot, color=blue, thick,
                 mark=*, mark options={fill=red, scale=1},text mark as node=true,
                 thick, mark size=2.0pt] coordinates {(1,42.52) (2, 44.64) (3,  48.62) (4,  48.60) (5, 48.14) (6,48.16) };
                 \draw [<->] [auto, thick,rounded corners=5pt, color=black] (-50,50) -- (80,50)node[above]{\tiny learning rate $=1.0^{-3}$} -- (100,50) ;
                 \draw [<->] [auto, thick,rounded corners=5pt, color=black] (100,150) -- (200,150)node[above]{\tiny learning rate $=1.0^{-4}$} -- (300,150) ;
                 \draw [<->] [auto, thick,rounded corners=5pt, color=black] (300,300) -- (400,300)node[above]{\tiny learning rate $=1.0^{-5}$} -- (500,300) ;

                \nextgroupplot[title={1500 training samples per class}]
                    \addplot[ybar, color=brown, fill=yellow!80!white!45, ybar legend] coordinates {(1, 43.05) (2, 45.88) (3,  48.70) (4, 48.82) (5, 49.14) (6, 49.00)  };
                    \addplot [sharp plot, color=blue, thick,
                 mark=*, mark options={fill=red, scale=1},text mark as node=true,
                 thick, mark size=2.0pt] coordinates {(1, 42.92) (2, 45.29)  (3, 49.10) (4, 48.59) (5, 49.25)(6, 49.59)};
                 \draw [<->] [auto, thick,rounded corners=5pt, color=black] (-50,50) -- (80,50)node[above]{\tiny learning rate $=1.0^{-3}$} -- (100,50) ;
                 \draw [<->] [auto, thick,rounded corners=5pt, color=black] (100,150) -- (200,150)node[above]{\tiny learning rate $=1.0^{-4}$} -- (300,150) ;
                 \draw [<->] [auto, thick,rounded corners=5pt, color=black] (300,300) -- (400,300)node[above]{\tiny learning rate $=1.0^{-5}$} -- (500,300) ;
            \end{groupplot}
        \end{tikzpicture}
        \ref{CombinedLegendBar}
        \caption{Top-1 Testing Accuracy of Places365: Our method with VGG16 Vs VGG-16}
        \label{PlusPlusCombinedBar}
    \end{figure*}
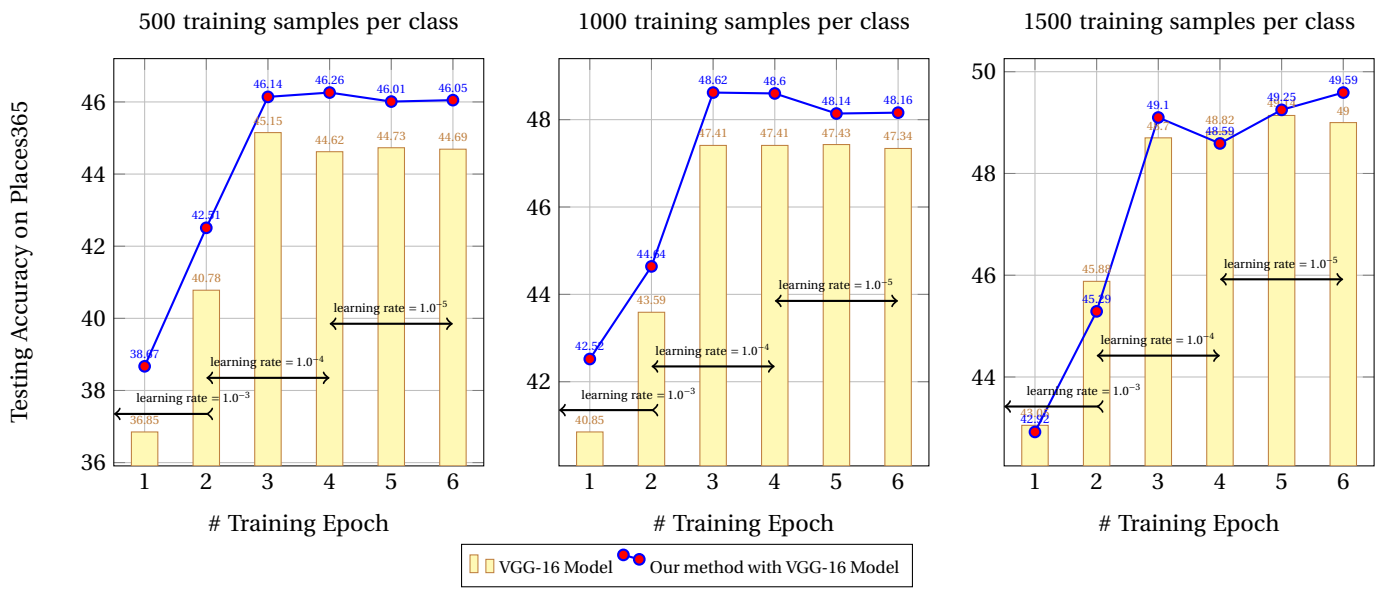

In the subsection, we train our method with four classic DCNN models including AlexNet, VGG16, 50-layer ResNet and Inception-GoogleNet. The comparison results on Scene15 and CIFAR10/100 are shown in Table 3-4 and Figure 5.

In Table 3, we include results from complex approaches that incorporate many cues and learning-optimal feature combinations and leading alternate approaches. For example, Zhou \emph{et al}. \cite{Zhounips2014} \cite{Zhoubolei16} term a new dataset (Places dataset), which almost contains more than 7 million images from 205 or 365 place categories, making it the largest image database of scenes and places so far. After pre-trained the large-scale dataset, around $92\%$ accuracy is obtained by \cite{Zhounips2014}, which is approaching nearly human-level performance at $95\%$. However, with the same model used in \cite{Zhounips2014}, our results is $94.8\%$, \textbf{which almost equal human-level performance}.

In Table 4, we take almost all the recent leading methods as rivals to evaluate our method, including ResNet, Inception, VGG-16, All Convnet Net \cite{ICLRJost}, Densely Net  \cite{Huang_2017_CVPR}, etc. It is easy to notice that our method with AlexNet, GoogleNet, and VGG16 model outperforms the existing state-of-the-art consistently on the three datasets.

\subsection{Comparison performance with the same DCNN architecture}

In the subsection, we involve almost all the recent well-known DCNN models to show the comparative performance between BP-based learning strategy and our method. Hence except for learning strategy, all the other experimental settings, including the learning rate, the monument rate, the batch-size, \textbf{the network architecture}, etc., are maintained the same. Then we train these DCNN models including AlexNet, VGG-16 Net, Google inception, DenseNet, and 50-layer ResNet with both BP-based method and our proposed method. The comparison results on the CIFAR10/100, the SUN397, and the Places365 datasets are shown in Table 5 and Figure 6-8.

\subsubsection{Comparison performance on Scene15, CIFAR10/100, and SUN397}

The results shown in Fig.6-8 and Table 5 indicate that DCNN models with our method significantly boost the learning capacity. As seen from Fig.6-8, it is easy to be noticed that with the same DCNN architecture, our proposed method generally provide better performance than that with BP-method.

In Table 5, we carry out a series of experiments under the two training environmental conditions (training from scratch, and training from a pretrained model) to evaluate the comparative performance. The advantage is obvious. For CIFAR10/100 and SUN397, the top-1 accuracies are close to 1$\%$ to 3$\%$ higher than the same DCNN model with BP method. Although the 1 to 3 percent top-1 accuracy boost seems to be a little improvement, it is not easy to obtain such improvements at the current stage. Let's take three well-known DCNN models for example. The VGG-16 that is an ILSVRC winner in the year of 2014 obtains 95.2 $\%$, 79.4$\%$, and 53.1 $\%$ top-1 accuracy on the three databases CIFAR10/100 and SUN397, respectively. However, for CIFAR10/100 datasets, the 2015 winner ResNet only provides 0.4 and 1.2 percent boost. Moreover, for SUN397, ResNet even provides a 1 percent lower compared to VGG-16.

Under the condition of training from scratch, 40-layer DenseNet that is proposed in 2017 also provides 0.2 percent accuracy boost compare to the performance of 44-layer ResNet on CIFAR10 dataset. However, unlike the above mentioned DCNN models which try to obtain performance improvements through a way of network structure optimization, our method never touch any network structure re-design task but achieves 1 to 3 percent boost by a non-iterative learning algorithm, which provides another direction in future research to further improve generalization performance of DCNN models.

\subsubsection{Comparison performance on Places365 and ImageNet}

\begin{table*}[!htb]
\centering
\caption{Classification accuracy on Scene-centric databases for the deep features of Object-centric databases (ImageNet). All the accuracy is the top-1 accuracy.}
\begin{tabular}{lccc}
\toprule
Method & Dataset &training image per category &Top-1 Accuracy  \\
\toprule
 \emph{Training from the ImageNet pretrained model} & \\
\,\,\,\,\,\,\,\,\,\,AlexNet, ImageNet-pretrained model  & Place365 & 1500  &40.13  \\
 \,\,\,\,\,\,\,\,\,\,Ours, ImageNet-pretrained AlexNet model &Place365 & 1500 &\textbf{42.21}  \\
 \,\,\,\,\,\,\,\,\,\,VGG-16,  ImageNet-pretrained model & Place365 & 1500 &49.00 \\
  \,\,\,\,\,\,\,\,\,\,Ours, ImageNet-pretrained VGG-16 model &Place365 & 1500 &\textbf{49.59} \\
 \midrule
  \emph{Training from scratch} & \\
  \,\,\,\,\,\,\,\,\,\,121-layer DenseNet  &ImageNet Mini & 200 &48.83  \\
\,\,\,\,\,\,\,\,\,\,Ours with 121-layer DenseNet & ImageNet Mini & 200 &\textbf{51.45}  \\
\,\,\,\,\,\,\,\,\,\,121-layer DenseNet  &ImageNet & 732-1300 &74.98\cite{Huang_2017_CVPR}  \\
\,\,\,\,\,\,\,\,\,\,Ours with 121-layer DenseNet & ImageNet & 732-1300 &\textbf{75.91}  \\

\bottomrule
\end{tabular}
\label{Table6}
\end{table*}

\begin{figure*}[!htb]
  \begin{center}
  \centering
\begin{tikzpicture}
\begin{axis}[
legend style={at={(1,1.1)},text=black, font=\fontsize{6}{7}\selectfont},
grid=both,
ybar,
                    width=15cm,
                    height=8cm,
enlargelimits=0.2,
ylabel={Training time per one training epoch (second)},
symbolic x coords={CIFAR10,CIFAR100,SUN397},
xtick=data,
nodes near coords,
nodes near coords align={vertical},
]
\addplot[ybar, color=red, fill=yellow!80!white!45, ybar legend] coordinates {(CIFAR10,581 ) (CIFAR100,587) (SUN397,336)};
\addplot[red, fill=red!80!white!80] coordinates {(CIFAR10,16) (CIFAR100,16) (SUN397,17)};
\addplot[ybar, color=blue, fill=blue!80!white!45, ybar legend] coordinates {(CIFAR10,123 ) (CIFAR100,121) (SUN397,387)};
\addplot[green, fill=green!80!white!80] coordinates {(CIFAR10,7) (CIFAR100,8) (SUN397,6)};
\legend{Original VGG-16, Ours method for recalculating fully-connected layers in VGG-16, Our method with VGG-16, Original DenseNet, Ours method for recalculating fully-connected layers in DenseNet, Our method with DenseNet}

\end{axis} \end{tikzpicture}

  \end{center}
  \caption{Training time per training epoch on Keras environment (one 1080 TI GPU): Our method vs original DCNN models. For DenseNet, we use 40-layer DenseNet on CIFAR10/100, and use 121-layer DenseNet on SUN397.}
  \label{fig:fom}
\end{figure*}
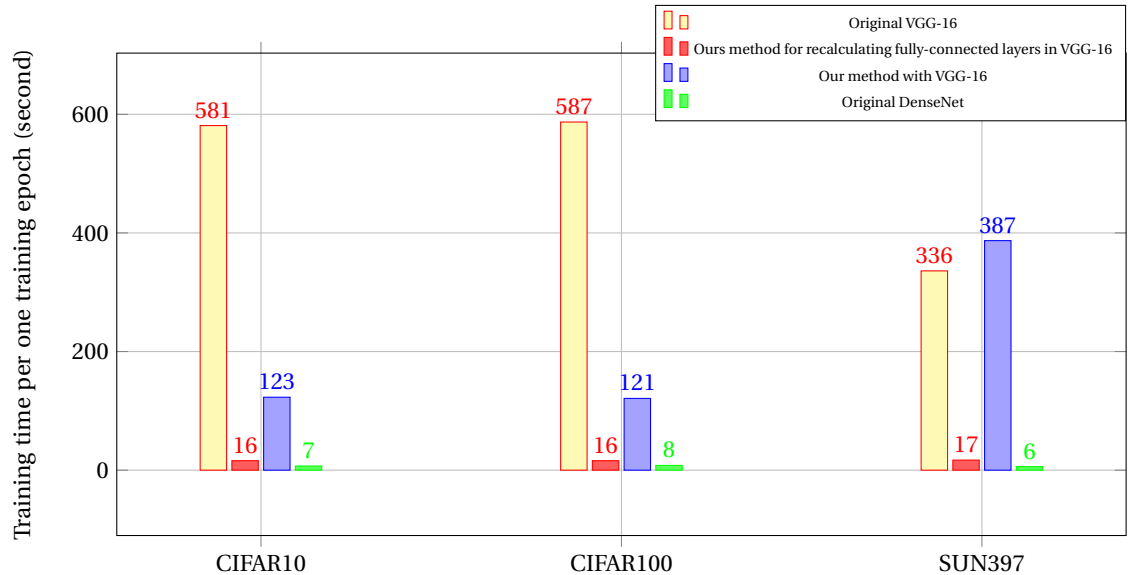

As far as we know, Places365 and ImageNet could be the largest datasets in image recognition task area. To further test the performance of our method on large-scale datasets, we select the recent large-scale datasets Places365 and ImageNet to evaluate our method. Similar to above-mentioned experiments, here we also try to cover recent well-known DCNN models under two training conditions that are training from a pretrained model or training from scratch model. The experimental results are shown in Fig.9-10 and Table 6.

First, we use ImageNet pretrained DCNN models as initial neuron parameters to evaluate the comparative performance of Places365 dataset. It is because ImageNet and Places365 are two different databases, the specialty of the units in the object-centric DCNN (ImageNet) and scene-centric DCNN (Places365) yield very different performances of generic visual features on a variety of recognition benchmarks. We report the top-1 accuracy of both our method and other two well-known DCNN models (AlexNet and VGG16) on Places365. As seen from Table 6 and figure 9-8, the results are quite similar as we mentioned in above subsection that our method has significant benefit regarding both learning effectiveness and generalization performance. As shown from figure 7, for the Alexnet with 182,500 training images (1500 training images per class), it needs six training epoches to provide 40.13$\%$ accuracy, while our method with two training epoches could provide 40.49$\%$ accuracy. Similar trends can be observed from Fig.8, the VGG-16 with our proposed method could also provide better performance than that of VGG-16 with BP method.

Second, we evaluate the comparative performance by training a scratch DenseNet model with both BP-method and our proposed method on ImageNet. Currently, our proposed method on Karas platform is only available for Single GPU environment. Thus to obtain the top-1 accuracy as fast as possible, here we select 121-layer DenseNet model because DenseNet has a relatively smaller number of parameters compared to VGG-16, AlexNet, or ResNet models. After around 17 days training, we finally obtain this top-1 accuracy of ImageNet.  As seen from Table V, the 121-layer DenseNet with our method provide around 1 percent top-1 accuracy boost compared to that of original 121-layer DenseNet.

\subsection{Computational Cost}

The proposed method is unable to shorten the training speeds in each learning iteration if network structure/size remains. However, our method does not add many computational workloads into the existing structure. Fig.11 shows that  compared to other DCNN models with iterative methods, the recomputation operation only bring a little extra computational workload. For example, the proposed method only uses 6-7 second to complete the recalculation operation in the three FC layers of VGG-16. For DenseNet model, the proposed non-iterative method even uses only 1 second to finish the recalculation operation in the one FC layers of DenseNet.

\section{Conclusion}

This paper introduces a new non-iterative learning strategy that replaces the iterative backpropagation used in conventional deep learning to update the parameters of fully-connected layers resulting in efficiency improvements in the training stage and performance boost in the testing accuracy. The experimental results demonstrate that the proposed model achieves the state-of-the-art results across several benchmark datasets compared to highly ranked object recognition methods.

However, the current model faces a limitation on the multi-GPU environment, and we consider to solve it in our future work.

{\small
\bibliographystyle{ieee}
\bibliography{reference}
}

\end{document}